\documentclass{article}

\usepackage{PRIMEarxiv}

\usepackage[utf8]{inputenc} % allow utf-8 input
\usepackage[T1]{fontenc}    % use 8-bit T1 fonts
\usepackage{hyperref}       % hyperlinks
\usepackage{url}            % simple URL typesetting
\usepackage{booktabs}       % professional-quality tables
\usepackage{amsfonts}       % blackboard math symbols
\usepackage{nicefrac}       % compact symbols for 1/2, etc.
\usepackage{microtype}      % microtypography
\usepackage{lipsum}
\usepackage{fancyhdr}       % header
\usepackage{graphicx}       % graphics
\graphicspath{{media/}}     % organize your images and other figures under media/ folder

\usepackage{amsmath}
\usepackage{multirow}

\title{Linear Chain Transformation: Expanding Optimization Dynamics for Fine-Tuning Large Language Models
}

\author{
  Yulong Wang, Chang Zuo \\
  State Key Laboratory of Networking and Switching Technology \\ School of Computer Science (National Pilot Software Engineering School) \\
  Beijing University of Posts and Telecommunications \\ 
  Beijing, China \\
  \texttt{\{wyl,changzuo\}email@bupt.edu.cn} \\
   \And
  Yin Xuan, Hong Li \\
  Guizhou University \\
  Guizhou, China\\
  \texttt{mec.xxu22,gs.ywen22@gzu.edu.cn} \\
   \And
  Ni Wei \\
  Fudan University \\
  Shanghai, China\\
  \texttt{weini@fudan.edu.cn} \\
}

\begin{document}
\maketitle

\begin{abstract}
Fine-tuning large language models (LLMs) has become essential for adapting pretrained models to specific downstream tasks. In this paper, we propose \emph{Linear Chain Transformation (LinChain)}, a novel approach that introduces a sequence of linear transformations during fine-tuning to enrich optimization dynamics. By incorporating multiple linear transformations into the parameter update process, LinChain expands the effective rank of updates and enhances the model's ability to learn complex task-specific representations. We demonstrate that this method significantly improves the performance of LLM fine-tuning over state-of-the-art methods by providing more flexible optimization paths during training, while maintaining the inference efficiency of the resulting model. Our experiments on various benchmark tasks show that LinChain leads to better generalization, fewer learnable parameters, and improved task adaptation, making it a compelling strategy for LLM fine-tuning.
\end{abstract}

% keywords can be removed

\section{Introduction}
% big background
Large language models (LLMs), such as ChatGPT, Claude, and LLaMA, have achieved remarkable success across a wide variety of natural language processing (NLP) tasks, ranging from text generation to question answering and summarization. Their ability to learn rich representations from vast amounts of data has enabled significant advancements in language understanding and generation. However, as LLMs continue to grow in size, the computational cost associated with fine-tuning these models for specific tasks becomes increasingly prohibitive. The challenge lies in balancing the power of large models with the need for efficient adaptation, especially when deploying them in real-world scenarios where computational resources may be limited.

% existing works
As LLMs continue to grow in scale, with billions of parameters, fine-tuning these models for specific tasks has become computationally expensive. To address this, several \textit{Parameter-Efficient Fine-Tuning (PEFT)} methods have been introduced, focusing on updating only a small subset of parameters while maintaining task-specific performance. A prominent method in this area is \textit{Low-Rank Adaptation (LoRA)}~\cite{DBLP:conf/iclr/HuSWALWWC22}, which updates a frozen pre-trained model by introducing a low-rank decomposition to the parameter updates. LoRA decomposes the weight update \( \Delta W \) into two smaller matrices \( A \in \mathbb{R}^{d_1 \times r} \) and \( B \in \mathbb{R}^{r \times d_2} \), reducing the number of trainable parameters from \( d_1 \times d_2 \) to \( (d_1 + d_2)r \)~\cite{DBLP:conf/iclr/HuSWALWWC22}. This has been effective in reducing memory usage and training costs, while preserving task performance.

However, LoRA’s low-rank updates can limit the model’s expressiveness, particularly for tasks requiring more complex feature interactions. This has led to several \textit{LoRA variants}, each attempting to balance efficiency and flexibility. For example, \textit{Mixture-of-Subspaces LoRA (MoSLoRA)} decomposes the low-rank updates into multiple subspaces and uses a learnable mixer to fuse these subspaces more flexibly, improving performance with negligible extra parameters~\cite{DBLP:journals/corr/abs-2406-11909}. %\textit{ReLoRA}, on the other hand, employs a reset mechanism that aggregates multiple low-rank updates over time, effectively simulating high-rank updates during training, allowing for performance improvements while keeping parameter costs low \cite{ReLoRA}.

% Beyond LoRA and its variants, other PEFT methods like \textit{LoKr} (which uses Kronecker products), \textit{LoHa} (which applies Hadamard products), and \textit{FLoRA} (which employs Tucker decomposition) aim to enrich the parameter space by introducing new structures to the low-rank updates \cite{MoSLoRA}. Despite their differences, these methods share a common goal: to enhance the capacity of the model to adapt to specific tasks without the need for full parameter updates, preserving the computational efficiency of LoRA while increasing expressiveness.

These approaches highlight the growing interest in fine-tuning models in a way that balances performance with efficiency. Yet, despite their innovations, these methods remain fundamentally limited by their reliance on low-rank approximations, which can fail to capture the complexity of certain tasks.

% \textbf{Motivation}. 
While LoRA and MoSLoRA reduce the computational cost of fine-tuning by limiting the number of trainable parameters, they impose a constraint on the model’s representational power due to their fixed low-rank approximation. This constraint can hinder performance on tasks that require more complex feature interactions. A natural extension of these methods would be to increase the expressiveness of the fine-tuning process while preserving its computational efficiency.

The motivation behind our work is to explore how additional flexibility in parameter updates can improve model performance without sacrificing efficiency. In particular, we hypothesize that introducing a \textit{chain of linear transformations}—which we term \textit{Linear Chain Transformation (LinChain)}—can provide the model with a richer set of optimization paths, thus enhancing its ability to adapt to task-specific data. This increased flexibility can help overcome the limitations of existing low-rank approaches by expanding the space of possible transformations.

% \textbf{Challenge}. The key challenge in designing a more flexible fine-tuning method is balancing expressiveness with efficiency. Adding more transformations introduces additional parameters, which, if not managed carefully, can lead to overfitting or increased computational costs. Furthermore, ensuring the stability of the optimization process when introducing multiple layers of transformations is critical to maintaining the model’s performance. The challenge, then, is to design a fine-tuning method that retains the computational benefits of low-rank adaptation while expanding the model’s ability to capture complex task-specific patterns.

% \textbf{Contribution of the Paper}. 
In this paper, we propose \textit{LinChain}, a novel fine-tuning method that introduces a sequence of linear transformations to enhance the model’s expressiveness while maintaining efficiency. LinChain addresses the limitations of existing low-rank adaptation methods by allowing for a richer set of parameter updates without introducing non-linearity. 

The contributions of this paper are summarized as follows.
\begin{itemize} \item This study uncovers a key insight: a chain of linear transformations enhances LLM training. The proposed LinChain method increases the optimization dynamics of LLM fine-tuning, improving the performance of fine-tuned models on tasks requiring diverse or complex transformations.

\item An in-depth analysis reveals that LinChain significantly extends the practical capabilities of LoRA, enabling more flexible and expressive updates while maintaining the efficiency benefits of low-rank adaptations.

\item Our experiments demonstrate that LinChain outperforms state-of-the-art fine-tuning methods across various tasks, even with fewer parameters, leading to faster convergence and improved task adaptation. \end{itemize}

The remainder of this paper is organized as follows: Section~\ref{sec:preliminary} introduces the preliminaries and our motivation, Section~\ref{sec:method} outlines the proposed LinChain method, Section~\ref{sec:experiment} presents our experimental results, Section~\ref{sec:relatedwork} reviews related works, and Section~\ref{sec:conclude} concludes with a discussion of future directions.

\section{Preliminaries and Motivation}
\label{sec:preliminary}

\subsection{LoRA and Its Extensions}
LoRA~\cite{DBLP:conf/iclr/HuSWALWWC22} is widely recognized for its effectiveness in fine-tuning LLMs while minimizing computational overhead. In LoRA, the weight update matrix \( \Delta W \) is decomposed into two low-rank matrices:
\begin{align}
\Delta W = A B^T, \quad A \in \mathbb{R}^{d_1 \times r}, \quad B \in \mathbb{R}^{r \times d_2},
\end{align}
where \( A \) and \( B \) are learned during fine-tuning, significantly reducing the number of trainable parameters. This structure allows for efficient adaptation to specific tasks by focusing on low-rank subspace updates.

However, LoRA’s fixed low-rank structure may not fully capture the complexity of tasks that require higher-dimensional or multi-modal representations. To address this, MoSLoRA~\cite{DBLP:journals/corr/abs-2406-11909}  introduces an extension by adding a learnable \textit{mixer matrix} between the LoRA matrices \( A \) and \( B \). The update equation in MoSLoRA is modified as:
\begin{align}
\Delta W = A W_m B^T,
\end{align}
where \( W_m \in \mathbb{R}^{r \times r} \) serves as a mixer matrix. MoSLoRA interprets this mixer in terms of the \textit{Mixture of Experts (MoE)} framework~\cite{DBLP:conf/iclr/ZadouriUAELH24}, where the mixer plays the role of a gating network. It helps match subspaces between the matrices \( A \) and \( B \). This addition of the mixer matrix allows MoSLoRA to capture more complex interactions between subspaces, resulting in improved task adaptation and performance.

\subsection{Motivation}
MoSLoRA enhances the flexibility and expressiveness of LoRA by introducing a mixer matrix, drawing on the MoE concept. However, we interpret the added mixer matrix differently. Rather than viewing the mixer matrix as a gating mechanism, we believe that the mixer matrix essentially adds an additional layer of linear transformation, which contributes to its improved performance over LoRA.

While it is true that in mathematics, a sequence of matrices can be collapsed into a single matrix through multiplication, their impact in gradient-descent-based optimization is fundamentally different. Inspired by the success of chain-of-thought prompting in large language models, where complex problems are broken down into simpler steps, we propose to decompose a single linear transformation into a chain of simpler matrices. The intuition is that the learnable parameters associated with each matrix in the chain are easier to optimize than those in a single, more complex matrix.

For example, a matrix that simultaneously performs shear and rotation can be split into two matrices: one handling shear and the other handling rotation. Each of these individual transformations is simpler than the combined one. Furthermore, introducing multiple matrices in a chain creates more paths for the optimization process. For instance, shear and rotation can be applied in different orders (shear followed by rotation, or vice versa), providing more flexibility for optimization. This multiplicity of optimization paths results in a richer search space for gradient descent to explore, increasing the chances of finding better-optimized parameter values.

By replacing the mixer in MoSLoRA with a sequence of matrices, we aim to further enhance the flexibility and expressiveness of the fine-tuning process while retaining the efficiency of low-rank approximations. This approach provides a way to address the limitations of fixed low-rank structures, especially in tasks that demand more nuanced and complex parameter updates.

% \subsection{Chain of Linear Transformations}
% To address the limitations of LoRA's low-rank structure, we propose \textit{Linear Chain Transformation (LinChain)}, which introduces a sequence of linear transformations that expands the capacity of the model to adapt to more complex tasks. In LinChain, the weight update is modified to include multiple linear transformations, represented as:
% \[
% \Delta W = A W_1 W_2 \cdots W_n B^T,
% \]
% where \( W_1, W_2, \dots, W_n \) are additional linear transformation matrices, with each matrix adding flexibility to the overall parameter update. The inclusion of these additional transformations increases the effective rank of the updates, enabling the model to represent more intricate relationships between tokens and task-specific features, while maintaining computational efficiency.

% By introducing this chain of linear transformations, LinChain retains the advantages of LoRA's efficiency but enhances the expressiveness of the fine-tuning process. This allows the model to better capture complex interactions in the data, improving performance across a wider range of tasks. Moreover, since the transformations are still linear, LinChain preserves the computational benefits of parameter-efficient methods, avoiding the overhead of non-linearities while providing a richer set of optimization paths.

\section{Linear Chain Transformation (LinChain)}
\label{sec:method}
\subsection{The Core Idea}
The main idea behind the proposed \textit{LinChain} finetuning method is to enhance the expressiveness of low-rank fine-tuning methods, such as LoRA, by introducing a series of linear transformations between the projection matrices. Instead of limiting the parameter update to a single low-rank transformation, LinChain employs multiple intermediate transformations to capture more complex relationships in the model's parameter space.

\begin{figure}
    \centering
    \includegraphics[width=0.5\linewidth]{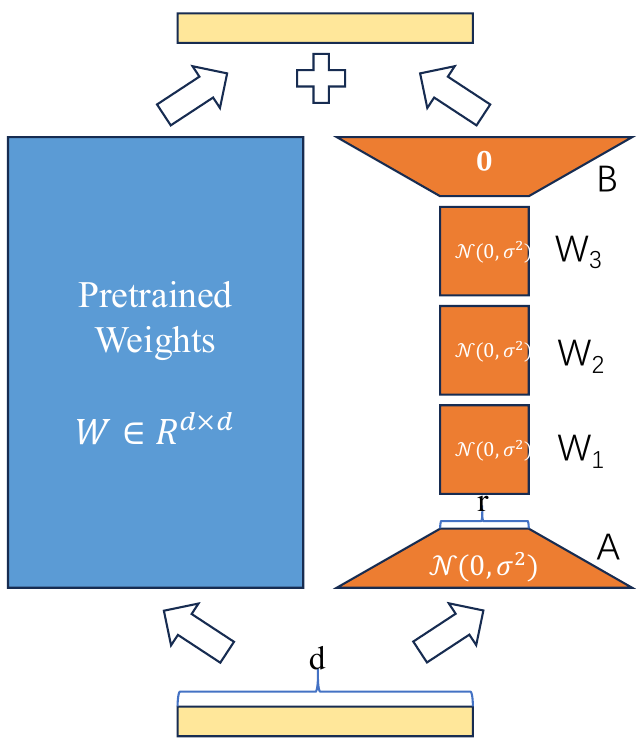}
\caption{Reparametrization in LinChain: $W_i$ ($i=1,2,3$) are trained alongside matrices $A$ and $B$.}
    \label{fig:overalldesign}
\end{figure}

In LoRA, the weight update is typically modeled as:
\begin{align}
\Delta W = A B^T,
\end{align}
where \( A \in \mathbb{R}^{d_1 \times r} \) and \( B \in \mathbb{R}^{r \times d_2} \) are the low-rank matrices learned during fine-tuning. While this is efficient, the expressiveness of this update is limited by the fixed low-rank structure.

To address this, LinChain introduces multiple intermediate transformations between \( A \) and \( B \), allowing for more flexibility in the parameter updates. Specifically, LinChain models the update as:
\begin{align}
\Delta W = A W_1 W_2 \cdots W_n B^T,
\end{align}
where \( W_1, W_2, \dots, W_n \) are learnable linear transformation matrices. Each \( W_i \) introduces an additional transformation layer, enriching the update process. The reparametrization in LinChain is illustrated in Fig.~\ref{fig:overalldesign}. We initialize \( A \) and \( W_i \) using a Kaiming uniform distribution~\cite{DBLP:conf/iccv/HeZRS15}, and set \( B \) as a zero matrix. This chain of linear transformations allows the model to more easily capture complex patterns and relationships, enhancing its ability to adapt to specific tasks.

LinChain remains computationally efficient because the added transformations are still linear, and the overall number of trainable parameters remains low. This approach preserves the benefits of low-rank adaptation while expanding the optimization dynamics through multilayer transformations.

\subsection{Analysis of Optimization Traces}

In LinChain, the chain of matrices introduces additional linear transformations between \( A \) and \( B \). Each matrix \( W_i \) in the chain adds an intermediate transformation, increasing the depth of the computational graph. This depth affects the optimization dynamics by introducing more paths through which gradients can propagate. In this analysis, we will quantify how the chain of matrices affects the number of optimization traces during gradient descent.

\paragraph{Definition of Optimization Traces}
An \textit{optimization trace} refers to the sequence of parameter updates during training, influenced by the gradients flowing through the computational graph. The number of optimization traces can be associated with the number of unique paths through which gradients can propagate from the loss function to each parameter.

\paragraph{Model Setup}
In LinChain, the adapted weight matrix is defined as:
\begin{align}
W_{\text{merge}} = W_0 + A W_1 W_2 \ldots W_n B,
\end{align}
where \( W_0 \in \mathbb{R}^{d_{\text{in}} \times d_{\text{out}}} \) is the frozen base weight matrix. \( A \in \mathbb{R}^{d_{\text{in}} \times r} \) and \( B \in \mathbb{R}^{r \times d_{\text{out}}} \) are low-rank adaptation matrices. \( W_i \) for \( i = 1, 2, \ldots, n \) are the mixer matrices forming the chain.

\paragraph{Forward Pass}
For an input \( x \in \mathbb{R}^{d_{\text{in}}} \), the output is:
\begin{align}
y = x W_{\text{merge}} = x W_0 + x A W_{\text{chain}} B,
\end{align}
where \( W_{\text{chain}} = W_1 W_2 \ldots W_n \).

\paragraph{Backward Pass and Gradient Computation}
To analyze the optimization traces, we compute the gradients of the loss \( L \) with respect to each parameter.

Let \( \Delta = \frac{\partial L}{\partial y} \in \mathbb{R}^{d_{\text{out}}} \). The gradients are:

1. Gradient with respect to \( A \):
   \begin{align}
   \frac{\partial L}{\partial A} = \left( x^T \Delta \right) B^T W_{\text{chain}}^T.
   \end{align}

2. Gradient with respect to \( B \):
   \begin{align}
   \frac{\partial L}{\partial B} = W_{\text{chain}}^T A^T x^T \Delta.
   \end{align}

3. Gradient with respect to each \( W_i \):
   \begin{align}
   \frac{\partial L}{\partial W_i} = \left( A^T x^T \Delta B^T \right) \cdot \frac{\partial W_{\text{chain}}}{\partial W_i}.
   \end{align}

The term \( \frac{\partial W_{\text{chain}}}{\partial W_i} \) involves products of all \( W_j \) for \( j \neq i \):
\begin{align}
\frac{\partial W_{\text{chain}}}{\partial W_i} = W_n W_{n-1} \ldots W_{i+1} \cdot I \cdot W_{i-1} \ldots W_1 = W_{\text{post}} \cdot I \cdot W_{\text{pre}},
\end{align}
where \( W_{\text{pre}} = W_{i-1} \ldots W_1 \) and \( W_{\text{post}} = W_n \ldots W_{i+1} \).

The gradient becomes:
\begin{align}
\frac{\partial L}{\partial W_i} = A^T x^T \Delta B^T W_{\text{post}}^T W_{\text{pre}}^T.
\end{align}

\paragraph{Number of Optimization Traces}
The number of unique paths through which gradients can flow from the loss \( L \) to each parameter is determined by the structure of the computational graph. For each \( W_i \), the gradient depends on all other \( W_j \), resulting in multiple paths.

For \( n \) matrices, the total number of optimization traces grows exponentially due to the combinatorial interactions between the matrices.

% ### 3. Parameter Efficiency
Despite the added flexibility, LinChain maintains the parameter efficiency of LoRA. The total number of trainable parameters in LinChain is given by:
\begin{align}
\text{Parameters in LinChain} = (d_1 + d_2)r + n r^2,
\end{align}
where \( n \) is the number of intermediate transformation matrices. The term \( n r^2 \) introduces a negligible number of additional parameters because \( n < r \ll \min(d_1, d_2) \), making LinChain a highly parameter-efficient approach.

\section{Experiments}
\label{sec:experiment}
In this section, we evaluate the performance of \textit{LinChain} through a series of experiments on standard NLP benchmarks. We compare LinChain to LoRA and its variants in terms of task performance, convergence rates, memory usage, training time, and computational cost.

\subsection{Benchmarking on NLP Tasks}

To assess the performance of LinChain, we conduct experiments on several widely-used NLP tasks, including commonsense reasoning, arithmetic reasoning, and natural language understanding. We compare LinChain to LoRA and MoSLoRA across these benchmarks. %, LoRA+, LoRA$^2$ and other fine-tuning methods across these benchmarks.

\paragraph{Datasets}

We evaluate the methods on the following datasets:

\begin{itemize}
    \item \textbf{Commonsense Reasoning:}  
    Our model is fine-tuned on COMMONSENSE170K, a combined training dataset consisting of 170,420 questions from multiple commonsense reasoning tasks. We evaluate the model on the following datasets:
    \begin{itemize}
        \item \textbf{PIQA}: Tests physical commonsense reasoning, where the model chooses one of the provided actions based on a hypothetical scenario. The test set includes 1,830 questions.
        \item \textbf{SIQA}: Focuses on reasoning about people's actions and their social consequences. The test set includes 1,954 questions.
        \item \textbf{HellaSwag}: Requires the model to select the most appropriate sentence completion given a context. The test set contains 10,042 questions.
        \item \textbf{WinoGrande}: Inspired by the Winograd Schema Challenge, this dataset tests commonsense reasoning by asking the model to fill in a blank with binary options. The test set has 1,267 questions.
        \item \textbf{ARC Easy (ARC-e)}: A set of grade-school level multiple-choice science questions. The test set includes 1,172 questions.
        \item \textbf{ARC Challenge (ARC-c)}: A more difficult version of ARC-e, designed to challenge co-occurrence-based methods. The test set contains 2,376 questions.
        \item \textbf{OBQA}: A knowledge-intensive, open-book QA dataset requiring multi-hop reasoning. The test set includes 500 questions.
    \end{itemize}
    
    \item \textbf{Arithmetic Reasoning:}  
    We fine-tune our model on MATH10K, a combined training dataset containing 9,919 problems from four tasks (GSM8K, MAWPS, MAWPS-single, and AQuA). Evaluation is performed on the following datasets:
    \begin{itemize}
        \item \textbf{AddSub}: Involves arithmetic word problems that require addition and subtraction. The test set contains 394 problems.
        \item \textbf{AQuA}: Contains algebraic word problems presented in multiple-choice format. The test set includes 253 problems.
        \item \textbf{GSM8K}: Comprises grade-school level math word problems that require multi-step reasoning. The test set includes 1,318 problems.
        \item \textbf{SVAMP}: Enhances the original Math World Problem (MWP) challenge by requiring robust reasoning invariant to structural alternations. The test set includes 999 problems.
        \item \textbf{MultiArith}: Consists of multi-step arithmetic problems. The test set contains 599 problems.
        \item \textbf{MAWPS}: Includes math word problems of varying complexity. The test set contains 237 problems.
        \item \textbf{SingleEq}: Contains grade-school math word problems that can be represented as single equations of varying lengths. The test set has 507 problems.
    \end{itemize}
    
    \item \textbf{GLUE (General Language Understanding Evaluation)~\cite{DBLP:conf/iclr/WangSMHLB19}:}  
    This benchmark includes a diverse set of NLP tasks such as sentiment analysis, textual entailment, and natural language inference. We carry out experiment under the following datasets in GLUE. 
    \begin{itemize}
    \item \textbf{CoLA}: The Corpus of Linguistic Acceptability consists of English acceptability judgments. The task is to predict whether a sentence is grammatically acceptable. The test set includes 1,043 examples.
    \item \textbf{SST-2}: The Stanford Sentiment Treebank includes sentences from movie reviews labeled as positive or negative sentiment. The test set contains 1,800 examples.
    \item \textbf{MRPC}: The Microsoft Research Paraphrase Corpus contains sentence pairs annotated with whether they are semantically equivalent. The test set includes 1,700 examples.
    % \item \textbf{QQP}: The Quora Question Pairs dataset consists of question pairs, with the task of determining whether they are semantically equivalent. The test set includes 391,000 examples.
    \item \textbf{STS-B}: The Semantic Textual Similarity Benchmark contains sentence pairs with human-annotated similarity scores from 1 to 5. The test set includes 1,400 examples.
    \item \textbf{MNLI}: The Multi-Genre Natural Language Inference Corpus involves predicting whether a hypothesis sentence is entailed by, contradicted by, or neutral to a premise sentence. The test set contains 9,800 examples.
    % \item \textbf{QNLI}: A dataset derived from the Stanford Question Answering Dataset (SQuAD), where the task is to determine whether a sentence contains the answer to a question. The test set contains 5,400 examples.
    \item \textbf{RTE}: The Recognizing Textual Entailment dataset contains sentence pairs annotated with entailment relationships. The test set includes 3,000 examples.
\end{itemize}

\end{itemize}

\paragraph{Evaluation Metrics}
We use the following metrics to evaluate performance:
\begin{itemize}
    \item \textbf{Accuracy}: Used for commonsense reasoning and arithmetic reasoning tasks, as well as for all datasets in GLUE except CoLA and STS-B.
    \item \textbf{Matthews Correlation Coefficient}: Used for the CoLA dataset in GLUE.
    \item \textbf{Pearson Correlation Coefficient}: Used for the STS-B dataset in GLUE.
\end{itemize}

\paragraph{Models} We use both a decoder-based model and an encoder-based model for our fine-tuning experiments: \begin{itemize} \item \textbf{LLaMA 3-8B Instruct}: A decoder-based large language model (LLM) developed by Meta AI, designed for a wide range of natural language understanding and generation tasks. It features 8 billion parameters and is fine-tuned on instruction-following data. \item \textbf{RoBERTa-base}: An encoder-based model with 12 layers, 768 hidden units, 12 attention heads, and 125 million parameters. \end{itemize} Both models are widely used in the literature for evaluating LLM fine-tuning methods.

\begin{table}[h]
\centering
\renewcommand{\arraystretch}{1.5}
\begin{tabular}{l|cccc}
\hline
\multirow{2}{*}{DataSet}  & LoRA & MoSLoRA & LinChain-3-16 (ours) & LinChain-2-8 (ours) \\
\cline{2-5}
&28.31 M   & 28.35 M & 28.43 M & 14.20 M  \\
\hline
ARC-c    & 74.7\%   & 79.0\%   & 80.5\%   & \textbf{81.6\%} \\
ARC-e    & 86.0\%   & 89.5\%   & \textbf{90.4\%}   & 90.0\% \\
OBQA     & 84.6\%   & 83.6\%   & \textbf{86.4\%}   & 85.0\% \\
PIQA     & 87.3\%   & 86.2\%   & \textbf{88.0\%}   & 85.5\% \\
SIQA     & 78.5\%   & 77.0\%   & \textbf{80.2\%}   & 78.8\% \\
WinoGrande   & 84.8\%   & 84.1\%   & \textbf{85.6\%}   & 84.8\% \\
HellaSwag  & 92.7\%   & 93.0\%   & 93.0\%   & \textbf{93.9\%} \\
Average     & 84.1\%   & 84.6\%   & \textbf{86.3\%}   & 85.7\% \\
\hline
\end{tabular}
\vspace{0.2cm}
\caption{Comparison of commonsense reasoning accuracy across different methods: LoRA, MoSLoRA, and LinChain. LoRA and MoSLoRA both utilize matrices $A$ and $B$ with a rank of 16. LinChain-3-16 introduces a chain of three $16 \times 16$ matrices while preserving the rank of $A$ and $B$ as 16. LinChain-2-8, on the other hand, incorporates a chain of two matrices—one sized $8 \times 16$ and the other $16 \times 8$—and uses matrices $A$ and $B$ with a rank of 8. LLaMA-8B-Instruct is employed as the pretrained model.}
\label{tab: commonsense}
\end{table}

\begin{table}[h]
\centering
\renewcommand{\arraystretch}{1.5}
\begin{tabular}{l|ccc}
\hline
DataSet     & LoRA  & MoSLoRA  & LinChain-3-16 (ours) \\
\hline
AQuA        & 28.0\%  & 29.9\%     & \textbf{30.7\%} \\
gsm8k       & 72.6\%  & 73.0\%     & \textbf{73.2\%} \\
SVAMP       & 78.1\%  & 80.2\%     & \textbf{81.8\%} \\
AddSub      & 89.9\%  & 92.2\%     & \textbf{92.9\%} \\
MultiArith  & 97.7\%  & 98.5\%     & \textbf{99.0\%} \\
\hline
Average     & 73.3\%  & 74.8\%     & \textbf{75.5\%} \\
\hline
\end{tabular}
\vspace{0.2cm}
\caption{Comparison of accuracy in arithmetic reasoning across different methods: LoRA, MoSLoRA, and LinChain. All methods utilize matrices $A$ and $B$ with a rank of 16.}
\label{tab:arithmatic}
\end{table}

\begin{table}[h!]
\centering
\renewcommand{\arraystretch}{1.5}
\begin{tabular}{lcccccc}%cc}
\hline
Model     & CoLA & MRPC & RTE  & STS-B & QNLI & SST-2 \\%& MNLI & QQP  \\
\hline
LoRA      & 59.7 & 88.7 & 75.3 & 90.3  & 92.6 & 93.9  \\%& 86.6 & 90.4 \\
LinChain-3-16  & \textbf{61.7} & 88.5 & \textbf{78.0} & \textbf{90.8}  & \textbf{92.8} & \textbf{94.2}    \\% & -    & -    \\
\hline
\end{tabular}
\vspace{0.2cm}
\caption{Comparison of LinChain and SOTA methods on the GLUE data collection. All methods utilize matrices $A$ and $B$ with a rank of 16.}
\label{tab:GLUE}
\end{table}

% \begin{table}[h]
% \centering
% \begin{tabular}{lccc}
% \hline
% \textbf{Model} & \textbf{GLUE (Avg. Acc)} & \textbf{SQuAD (F1/EM)} & \textbf{NER (F1)} \\
% \hline
% LoRA & 83.2\% & 87.4/80.9 & 91.1 \\
% MoSLoRA & 84.0\% & 88.2/82.0 & 92.3 \\
% LinChain (Ours) & \textbf{85.4\%} & \textbf{89.3/83.5} & \textbf{93.7} \\
% \hline
% \end{tabular}
% \caption{Performance of LinChain compared to LoRA and MoSLoRA on NLP benchmarks.}
% \label{tab:results}
% \end{table}

% LinChain demonstrates superior performance across all tasks, particularly for NER and SQuAD, which require capturing intricate relationships between tokens.

\subsection{Effectiveness of Chain-of-Linear-Transformations}
\label{sec:effectiveness}

One of the key contributions of LinChain is the introduction of multiple linear transformations, which provide additional flexibility in representing complex feature interactions. In this subsection, we analyze how this chain-of-transformations impacts task performance and convergence rates.

\paragraph{Performance Improvement}
The introduction of intermediate transformations significantly enhances the model's ability to capture high-dimensional dependencies. As shown in the fourth column of Table~\ref{tab: commonsense}, LinChain achieves a notable performance boost over both LoRA and MoSLoRA on commonsense reasoning tasks. With a comparable number of learnable parameters (28.43M for LinChain vs. 28.31M for LoRA and 28.35M for MoSLoRA), LinChain attains an average commonsense accuracy of 86.3\%, significantly outperforming LoRA (84.1\%) and MoSLoRA (84.6\%). Furthermore, as highlighted in the last column of Table~\ref{tab: commonsense}, even when using half the parameters (14.20M) by lowering the rank (8 vs. 16), LinChain still achieves an impressive average accuracy of 85.7\%, surpassing both LoRA and MoSLoRA. This improvement is particularly pronounced in more challenging datasets, such as ARC-c and HellaSwag.

We also assess the effectiveness of the proposed LinChain method on the more challenging arithmetic reasoning datasets, as shown in Table~\ref{tab:arithmatic}. The results in the last column demonstrate that LinChain outperforms both LoRA and MoSLoRA across all five datasets, achieving an average accuracy of 75.5\%.

Additionally, we evaluate the performance of LinChain on the GLUE benchmark. As shown in Table~\ref{tab:GLUE}, LinChain consistently surpasses LoRA on most datasets within GLUE, further showcasing its superior fine-tuning capabilities.

The effectiveness of LinChain can be attributed to the additional optimization traces introduced by the method. These traces allow the model to explore a broader range of optimization paths during gradient descent, leading to a better fit for task-specific data—especially for tasks with complex dependencies.

\begin{figure}[h]
\centering
\includegraphics[width=0.7\textwidth]{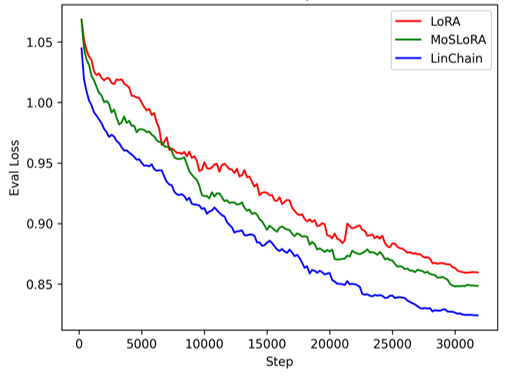}
\caption{Training loss curves for LinChain, LoRA, and MoSLoRA on the Commonsense170K dataset. LinChain demonstrates faster convergence and achieves a lower final loss. All methods employ matrices $A$ and $B$ with a rank of 16. In LinChain, three additional $16 \times 16$ matrices are inserted between matrices $A$ and $B$.}
\label{fig:convergence}
\end{figure}

\paragraph{Convergence Analysis}
In addition to performance improvements, LinChain also demonstrates faster convergence compared to LoRA and MoSLoRA. Figure~\ref{fig:convergence} shows the training loss curves for the different models on the Commonsense170K dataset.

LinChain converges more quickly, achieving lower final loss after fewer training epochs. This suggests that the chain-of-transformations facilitates more efficient optimization, likely due to the additional flexibility in the parameter updates provided by the intermediate matrices.

\subsection{Efficiency Evaluation}

While LinChain introduces additional parameters compared to LoRA, it remains computationally efficient. In this subsection, we compare LinChain to LoRA and MoSLoRA in terms of memory usage, training time, and computational cost.

\paragraph{Memory Usage}
The second column of Table~\ref{tab:efficiency} presents the memory usage of the different methods during training. Although LinChain introduces multiple intermediate matrices, the overall increase in memory usage is minimal with the same rank of 16. This is because the transformations are still low-rank and the number of parameters added by the chain (see the fourth column of Table~\ref{tab:efficiency}) is relatively small compared to full fine-tuning.

LinChain uses slightly more memory than LoRA and MoSLoRA, but it remains far more efficient than full fine-tuning.

\paragraph{Training Time}
In terms of training time, LinChain performs comparably to LoRA and MoSLoRA (see the third column of Table~\ref{tab:efficiency}). The additional matrix multiplications introduced by the chain result in a small increase in computation time per epoch, but this is offset by the faster convergence seen in Section~\ref{sec:effectiveness}. Overall, LinChain achieves better performance in a similar number of training epochs.

% \paragraph{Computational Cost}
% The computational cost of LinChain is determined by the number of transformation matrices in the chain, denoted \( n \), and the rank \( r \) of the adaptation matrices. As detailed in Section 3.2, the total number of parameters in LinChain is given by:
% \[
% \text{Parameters in LinChain} = (d_1 + d_2)r + n r^2.
% \]
% While the computational cost increases with \( n \), the increase remains manageable for most practical applications. By selecting appropriate values for \( n \) and \( r \), LinChain can be efficiently scaled to larger models and tasks, balancing expressiveness with efficiency.

\paragraph{Overall Efficiency}
Despite the introduction of additional parameters, LinChain offers a favorable trade-off between expressiveness and efficiency. The small increase in memory and training time is justified by the significant gains in task performance and faster convergence. This makes LinChain a viable alternative for tasks that require a higher degree of expressiveness, without incurring the full computational cost of conventional fine-tuning.

\begin{table}[h]
\centering
\begin{tabular}{lccc}
\hline
\textbf{Model} & \textbf{Memory Usage (GB)} & \textbf{Training Time (hours/epoch)} & \textbf{Parameters (M)} \\
\hline
LoRA & 24.49 & 2.83 & 28.31 \\
MoSLoRA & 24.49 & 2.97 & 28.35 \\
LinChain (Ours) & 24.50 & 3.13 & 28.43 \\
\hline
\end{tabular}
\vspace{0.2cm}
\caption{Memory usage, training time, and parameter count comparison of LinChain, LoRA, and MoSLoRA.}
\label{tab:efficiency}
\end{table}

\section{Conclusion}
\label{sec:conclude}
In this paper, we present \emph{LinChain}, a novel parameter-efficient fine-tuning method for LLMs that extends the low-rank adaptation framework by introducing a chain of linear transformations. Our proposed method addresses the limitations of existing techniques such as LoRA and MoSLoRA, which, despite their efficiency, struggle to capture complex task-specific feature interactions. By decomposing the weight update into a sequence of transformations, LinChain significantly improves the expressiveness and flexibility of the model without compromising computational efficiency.

Our experimental results demonstrate that LinChain outperforms both LoRA and its variants across a variety of tasks, including commonsense reasoning, arithmetic reasoning, and various NLP tasks. LinChain's ability to model high-order dependencies between features is particularly evident in tasks that require complex token-level interactions, where it achieves superior performance in terms of accuracy and correlation coefficient. Furthermore, LinChain shows faster convergence during training, thanks to its richer optimization paths facilitated by the chain of transformations.

In terms of efficiency, LinChain introduces only a minimal increase in memory usage and computational cost compared to LoRA and MoSLoRA. The flexibility offered by the additional transformation matrices leads to better task adaptation with only a slight trade-off in parameter efficiency. The overall balance between expressiveness and computational cost makes LinChain a highly effective solution for fine-tuning large models on complex tasks. We will conduct further experiments to validate the effectiveness of the proposed LinChain fine-tuning method.

\bibliographystyle{unsrt}  
\bibliography{references}

\end{document}